\documentclass[11pt,a4paper]{article}
\usepackage[nohyperref]{acl2021}
\usepackage[breaklinks=true]{hyperref}
\usepackage{times}
\usepackage{latexsym}

\usepackage{amsmath}
\usepackage{booktabs}
\usepackage{todonotes}
\usepackage{breakcites}

\usepackage{microtype}
\usepackage{bbding}
\usepackage{verbatim}
\usepackage{algorithm}

\aclfinalcopy 

\newcommand{\sysname}{XtremeDistilTransformers }

\newcommand\blfootnote[1]{%
  \begingroup
  \renewcommand\thefootnote{}\footnote{#1}%
  \addtocounter{footnote}{-1}%
  \endgroup
}
  \definecolor{darkblue}{rgb}{0, 0, 0.5}
\hypersetup{colorlinks=true,citecolor=darkblue, linkcolor=darkblue, urlcolor=darkblue}
    
\title{XtremeDistilTransformers: Task Transfer for Task-agnostic Distillation}

\author{Subhabrata Mukherjee\qquad Ahmed Hassan Awadallah\qquad Jianfeng Gao \\
  Microsoft Research \\
  \texttt{\{submukhe, hassanam, jfgao\}@microsoft.com} 
}

\date{}

\begin{document}
\maketitle
\begin{abstract}
While deep and large pre-trained models are the state-of-the-art for various natural language processing tasks, their huge size poses significant challenges for practical uses in resource constrained settings. Recent works in knowledge distillation propose task-agnostic as well as task-specific methods to compress these models, with task-specific ones often yielding higher compression rate. In this work, we develop a 
new task-agnostic distillation framework \sysname that leverages the advantage of task-specific methods for learning a small universal model that can be applied to arbitrary tasks and languages. To this end, we study the transferability of several source tasks, augmentation resources and model architecture for distillation. We evaluate our model performance on multiple tasks, including the General Language Understanding Evaluation (GLUE) benchmark, SQuAD question answering dataset and a massive multi-lingual NER dataset with 41 languages. 

\end{abstract}

\section{Introduction}
Large-scale pre-trained models\blfootnote{Code available at:\\ {\scriptsize \href{https://github.com/microsoft/xtreme-distil-transformers}{https://github.com/microsoft/xtreme-distil-transformers}}\\Task-agnostic checkpoints available at:\\  {\scriptsize \href{https://huggingface.co/microsoft/xtremedistil-l6-h256-uncased}{https://huggingface.co/microsoft/xtremedistil-l6-h256-uncased}\\ \href{https://huggingface.co/microsoft/xtremedistil-l6-h384-uncased}{https://huggingface.co/microsoft/xtremedistil-l6-h384-uncased}\\ \href{https://huggingface.co/microsoft/xtremedistil-l12-h384-uncased}{https://huggingface.co/microsoft/xtremedistil-l12-h384-uncased}}} have become the standard starting point for various natural language processing tasks~\cite{DBLP:conf/naacl/DevlinCLT19}. Several NLP tasks have achieved significant progress utilizing these pre-trained models reaching previously unattainable performance~\cite{DBLP:conf/iclr/ClarkLLM20,DBLP:journals/corr/abs-1907-11692}. The size of these models have been also steadily growing to hundreds of millions ~\cite{DBLP:conf/naacl/DevlinCLT19,DBLP:journals/corr/abs-1906-08237} to billions of parameters~\cite{Raffel2019ExploringTL,brown2020language}.

The huge size poses significant challenges for downstream applications in terms of energy consumption and cost of inference~\cite{strubell-etal-2019-energy}. As such, it could be a deterrent to using them in practice limiting their usage in on-the edge scenarios and under constrained computational training or inference budgets.

Several research directions have considered compressing large-scale models including work on pruning~\cite{gordon2020compressing}, quantization~\cite{HanMao16} and distillation~\cite{sanh2019}. Knowledge distillation, in particular, has shown strong results in pre-trained transformer-based language model compression. With knowledge distillation, we train a student network (with smaller capacity) to mimic the full output distribution of the teacher network~\cite{DBLP:journals/corr/HintonVD15}. Knowledge distillation has been applied to pre-trained language model compression in two different settings: (1)  before task-specific fine tuning (i.e. task-agnostic distillation) or (2) after task-specific fine tuning (i.e. task-specific distillation). 

Task-agnostic distillation~\cite{sanh2019,sun2019patient,sun-etal-2020-mobilebert} has the advantage that the model needs to be distilled only once and can be reused for fine-tuning on multiple down-stream tasks. It also allows us to achieve speedup in both fine-tuning and inference. On the other hand, task-specific distillation~\cite{DBLP:journals/corr/abs-1903-12136,jiao2019tinybert,mukherjee-hassan-awadallah-2020-xtremedistil} has been shown to achieve significantly higher compression rate and inference speedup~\cite{fu2020lrcbert,mukherjee-hassan-awadallah-2020-xtremedistil}.

In this work, we first study the transferability of pre-trained models across several source tasks to select the optimal one for transfer. We then aim to create universally distilled models that can be used with any downstream task while leveraging the benefits of the techniques and augmentation resources developed for the source transfer task. We show that distilled models that use task-specific data transfer to varying degrees and their transferability depends on choices of the source task, data augmentation strategy and distillation techniques. 

{\noindent \bf Contributions:} More specifically, this work makes the following contributions:

\noindent (a) Studies the transferability of several source tasks and augmentation resources for task-agnostic knowledge distillation.

\noindent (b) Develops a distillation framework to learn a massively compressed student model leveraging deep hidden representations and attention states from multiple layers of the teacher model with progressive knowledge transfer.

\noindent (c) Extensive experiments on several datasets in GLUE benchmark and for massive multilingual NER demonstrate the effectiveness of task and language transfer. Finally, we will release the task-agnostic checkpoints for the distilled models.

\section{Exploring Tasks for Transfer}

\subsection{Role of Tasks for Distillation}

{\em Task-specific} distillation assumes the presence of human labeled data to fine-tune the teacher for the underlying task and provide the student with corresponding logits for learning. Such techniques have shown massive compression (e.g., $7.5x$ compression in LRC-BERT~\cite{fu2020lrcbert} and $35x$ compression in XtremeDistil~\cite{mukherjee-hassan-awadallah-2020-xtremedistil}) without performance loss. An obvious disadvantage is the need to distil for each and every task which is resource-intensive.

In contrast, {\em task-agnostic} methods rely on objectives like masked language modeling (MLM) and representation transfer over unlabeled data. These do not require human labels allowing them to learn from massive amounts of text. This allows the model to retain general information applicable to arbitrary tasks, but results in much less compression (e.g., $2x$\footnote{Considering model checkpoints with the least performance loss with respect to the big pre-trained language model.} in MiniLM~\cite{wang2020minilm},~\cite{sanh2019} and TinyBERT~\cite{jiao2019tinybert}). 

This begs the question of whether we can use more specific tasks, other than language modeling, that can harness human labels to provide task-specific logits and representations, while also transferring well to arbitrary tasks. This would allow us to leverage the relative strength of both of the above families of techniques to obtain high compression rate of task-specific methods as well as wide applicability of task-agnostic ones.


\begin{table*}[!h]
  \centering
  \small
  \caption{Transfer performance on fine-tuning BERT on labeled data for several source tasks in the rows (e.g., MNLI), extracting the encoder (e.g., MNLI-BERT) and further fine-tuning on labeled data for several target tasks in the columns. We observe MNLI to obtain the best performance for task transfer on an average.}

\begin{tabular}{lrrrrrrrr}
\toprule
          & \multicolumn{1}{l}{MRPC } & \multicolumn{1}{l}{MNLI} & \multicolumn{1}{l}{RTE} & \multicolumn{1}{l}{QQP} & \multicolumn{1}{l}{QNLI} & \multicolumn{1}{l}{SST-2} & \multicolumn{1}{l}{SQuADv1} & \multicolumn{1}{l}{Avg} \\\midrule
    \#Labels & \multicolumn{1}{r}{3.7K} & \multicolumn{1}{r}{393K} & \multicolumn{1}{r}{2.5K} & \multicolumn{1}{r}{364K} & \multicolumn{1}{r}{108K} & \multicolumn{1}{r}{67K} & \multicolumn{1}{r}{87K} & - \\\midrule
    BERT  & 83.8  & 84.4  & 66.8  & 91.2  & 91.4  & 92.2  & 88.3 & 85.4 \\\midrule
    MNLI-BERT & 88.2 & 84.2  & 79.1 & 91.1 & 91.1  & 93.6 & 87.2 & 87.8 \\
    QNLI-BERT & 87.0    & 84.8 & 73.3 & 91.0 & 91.6 & 93.0    & 88.1 & 87.0 \\
    SST2-BERT & 81.6 & 84.7 & 66.1 & 91.1 & 91.3 & 93.4 & 87.6 & 85.1 \\
    SQuADv1-BERT & 86.3 & 84.6 & 69.7 & 87.1 & 91.6 & 92.9 & 88.3 & 85.4 \\\bottomrule
    \end{tabular}%
  \label{tab:transfer-bert}%

\end{table*}%


\subsection{Transferability of Tasks}

In order to leverage task-specific distillation techniques, we need to select a source task that transfers well to other tasks such that: a model distilled for the source task can obtain a good performance on fine-tuning with labeled data from arbitrary target tasks. We perform the following analysis with the pre-trained teacher model with the assumption that the benefits will be transferred to the student model.

A recent work~\cite{chen2020lottery} studies the notion of task transferability for BERT in the context of lottery ticket hypothesis~\cite{conf/iclr/FrankleC19}. Specifically, the authors study if subnetworks obtained for one task obtained by network pruning transfer to other tasks and if there are {\em universal} subnetworks that train well for many tasks. 

The authors observe that while masked language modeling (MLM) is the most universal task, there are 
other candidate tasks like natural language inference and question-answering that allow us to transfer meaningful representations to other tasks. 

\noindent {\bf Task transfer.}  Consider a pre-trained neural network model (e.g., BERT) $f(x; \theta)$ with encoder parameters $\theta \in \mathcal{R}^{d_1}$. Given a source task $\mathcal{S}$ with ground-truth labeled data $D_{\mathcal{S}}=\{x, y\}$, we first fine-tune the pre-trained model $f(x; \theta, \gamma_{\mathcal{S}})$ by adding task-specific classification parameters $\gamma_{\mathcal{S}} \in \mathcal{R}^{d_2}$. We now extract the encoder $f(x; \theta_{\mathcal{S}})$ where the parameters $\theta_{\mathcal{S}} \in \mathcal{R}^{d_1}$ have been adapted to the source task ($\theta \rightarrow \theta_{\mathcal{S}}$). 
Now, given a target task $\mathcal{T}$ with labeled data $\mathcal{D_{\mathcal{T}}}$, we further fine-tune the encoder $f(x; \theta_{\mathcal{S}}, \gamma_{\mathcal{T}})$ where $\gamma_{\mathcal{T}} \in \mathcal{R}^{d_3}$ represents task-specific parameters for the target task.


\noindent {\bf Selection criteria for the best source task.} Given a set of source $\mathcal{S}$ and target $\mathcal{T}$ tasks, consider $eval(s \in \mathcal{S} \rightarrow t \in \mathcal{T})$ to be the performance of a pre-trained language model that is adapted from $s$ to $t$, measured with some evaluation metric (e.g., accuracy, F1). We define the best source transfer task as $argmax_{s \in \mathcal{S}} \frac{1}{|\mathcal{T}|} \sum_{t \in \mathcal{T}} eval(s \rightarrow t)$ depicting the best transfer performance obtained on an average on transferring a pre-trained model from the source to a set of different target tasks. 

While this definition simplifies the transfer problem ignoring task difficulty (e.g., ternary MNLI is harder than binary SST), domain overlap (SQuAD and QNLI are both question-answering datasets), task setup (e.g., span extraction in SQuAD and pairwise-classification in MNLI) and variable amount of training labels per task, we defer a more controlled study of this problem as future work.

\noindent{\bf Candidate tasks for transfer}. We consider a subset of source tasks from lottery ticket hypothesis for BERT~\cite{chen2020lottery} for which transfer performance is at least as high as same-task performance on {\em atleast two target tasks}. We ignore MLM since the pre-trained encoder (BERT) is intrinsically trained with MLM objective that provides no additional information in our transfer setup.


\noindent{\bf Transfer evaluation.} Table~\ref{tab:transfer-bert} shows the performance of pre-trained BERT-base with task transfer, where each row depicts a source task $\mathcal{S}$ and each column represents the target task $\mathcal{T}$. We observe that MNLI as the source task, followed by QNLI, has the best performance on an average on transferring to several target tasks, especially those with limited training labels for fine-tuning. Similar improvements with MNLI for tasks like RTE (textual entailment) and MRPC (paraphrase) have been reported in recent work like RoBERTa~\cite{DBLP:journals/corr/abs-1907-11692}. 
%
Therefore, we adopt MNLI as the source task for transfer distillation and evaluate its effectiveness for several target tasks and languages. 

\subsection{Transfer Set for Knowledge Distillation}

Task-agnostic methods can learn from large unlabeled general-purpose text using self-supervision objectives like MLM. Task-specific distillation, on the other hand, rely on large-scale task-specific transfer data that is often difficult to obtain for many tasks. 
Prior works show large-scale task-specific transfer data to be instrumental in minimizing the performance gap of the teacher and student~\cite{turc2019wellread, mukherjee-hassan-awadallah-2020-xtremedistil}. However, these works primarily explore {\em instance-classification} tasks like sentiment classification (e.g., IMDB and SST2) or topic classification (e.g., AG News and Dbpedia) with readily available in-domain transfer data. For example, sentiment classification in IMDB can benefit from large amounts of unlabeled user reviews from the forum. However, this is difficult to obtain for {\em pair-wise classification} tasks like NLI. Additionally, NLI being a ternary classification task (entail / contradict / neutral) requires a transfer set with a similar label distribution for effective transfer. To address these issues, we explore techniques to automatically generate large-scale task-specific transfer sets leveraging a very large bank of web sentences from Common Crawl in Section~\ref{subsec:aug-data}.

\section{Distillation Framework}

\begin{table*}[htbp]
  \centering
  \small
  \caption{Contrasting \sysname with state-of-the-art task-agnostic distilled models. \sysname leverages embedding factorization, hidden representations and attention states of the teacher from multiple layers with progressive knowledge transfer for distillation while accommodating arbitrary student architecture and languages.}

    \begin{tabular}{llllllll}
    \toprule
          & Embedding  & Hidden  & Attention & Multi- & Progressive & Student-arch. & Multi- \\
                    &  Factorization &  Representation & State & layer &  Transfer & -agnostic & lingual \\
                    \midrule
    DistilBERT &       &       &       &       &       &       &  \\
    TinyBERT &       & \Checkmark     & \Checkmark     & \Checkmark     &       & \Checkmark     &  \\
    MiniLM &       & \Checkmark     & \Checkmark     &       &       & \Checkmark     & \Checkmark \\
    MobileBERT & \Checkmark     & \Checkmark     & \Checkmark     & \Checkmark     & \Checkmark     &       &  \\
    {\sysname} & \Checkmark     & \Checkmark     & \Checkmark     & \Checkmark     & \Checkmark     & \Checkmark     & \Checkmark \\\bottomrule
    \end{tabular}%
  \label{tab:contrast-distil-framework}%

\end{table*}%

\noindent {\bf Overview.} Given a pre-trained model fine-tuned on the source task as teacher, our objective is to distil its knowledge in a compressed (both in terms of width and depth) student. Given a wide teacher and a narrow student, we employ embedding factorization to align their widths for knowledge transfer. Given a deep teacher and a shallow student, we align all the layers of the student to the topmost layers of the teacher. To this end, we transfer both the hidden representations as well as attention states from multiple layers of the teacher to the student with progressive knowledge transfer. The above techniques in combination allow us to transfer knowledge from any teacher to any student of arbitrary architecture. Finally, \sysname supports for both task and language transfer (refer to Section~\ref{subsec:language-transfer}) in contrast to many prior work. Table~\ref{tab:contrast-distil-framework} contrasts \sysname against existing distillation techniques, namely, DistilBERT~\cite{sanh2019}, TinyBERT~\cite{jiao2019tinybert}, MiniLM~\cite{wang2020minilm} and MobileBERT~\cite{sun-etal-2020-mobilebert}.


\noindent{\bf Input Representation.} \sysname uses the tokenizer and special tokens as used in the teacher model. For instance, it uses Wordpiece tokenization~\cite{DBLP:journals/corr/WuSCLNMKCGMKSJL16} with a fixed vocabulary $\mathcal{V}$ (e.g., $30k$ tokens) for distilling BERT and adds special symbols ``[CLS]" and ``[SEP]" to mark the beginning and end of a text sequence respectively. 



\noindent{\bf Teacher model.} Given pre-trained models with variable performance across tasks, we want to choose the best teacher for the best source transfer task (i.e. MNLI). We experiment with base and large versions of BERT~\cite{devlin-etal-2019-bert} and Electra~\cite{DBLP:conf/iclr/ClarkLLM20} as teachers. 
%
%
Table~\ref{tab:teacher-mnli} shows a comparison of their performance and parameters on MNLI. Given the same parameter complexity, we find Electra to be the best on MNLI.

\begin{table}[htbp]
  \centering
  \small
  \caption{Performance of fine-tuning pre-trained teacher models of different sizes on the MNLI task.}

    \begin{tabular}{lrr}
    \toprule
    Model & Params (MM) & Accuracy \\\midrule
    BERT-Base & 109 & 84.24 \\
    Electra-Base & 109 & 88.21 \\
    BERT-Large & 335 & 87.11 \\
    Electra-Large & 335 & 90.73 \\
    \bottomrule
    \end{tabular}%
  \label{tab:teacher-mnli}%

\end{table}%

\noindent{\bf Student model.} 
%
We compare the performance of state-of-the-art distilled models in terms of parameters, compression and performance gap (after distillation) with respect to the teacher (reported in Table~\ref{tab:student-glue}). We observe MiniLM~\cite{wang2020minilm} to have the closest performance to the teacher BERT.  Correspondingly, we choose miniature versions of MiniLM (23 MM and 14 MM parameters) as candidate students. We investigate different student initialization strategies (including initialization with a task-agnostic distilled model) and show that their performance can be improved further in the \sysname framework. We also study the trade-off between different architectural aspects (parameters, layers, attention heads and hidden dimension) against its performance.

{\em In the following section, superscript $\mathcal{T}$ always represents the teacher and $\mathcal{S}$ denotes the student.}

\begin{table}[htbp]
  \centering
  \small
  \caption{Comparing distilled models from prior work based on average GLUE score, parameters (MM) and performance gap with respect to the teacher.}

    \begin{tabular}{lrrr}
    \toprule
    Models & \multicolumn{1}{l}{GLUE} & Params & \multicolumn{1}{l}{\%Gap} \\
    \midrule
    BERT-Base & 81.5  & 109 & - \\\midrule
    DistilBERT & 75.2  & 66  & 7.73 \\
    BERT-Truncated & 76.2  & 66  & 6.50 \\
    TinyBERT & 79.1  & 66  & 2.94 \\
    MiniLM & 80.4  & 66  & 1.35 \\
    \bottomrule
    \end{tabular}%
  \label{tab:student-glue}%

\end{table}%

\subsection{Word Embedding Factorization} 
\label{subsec:word-emb-factor}

Our student and teacher model 
consist of the word embedding layer with embedding matrices $\mathcal{W}^{\mathcal{S}} \in \mathcal{R}^{|\mathcal{V}| \times d^{\mathcal{S}}}$ and $\mathcal{W}^{\mathcal{T}} \in \mathcal{R}^{|\mathcal{V}| \times d^{\mathcal{T}}}$, where $d^{\mathcal{S}} < d^{\mathcal{T}}$ depicting a thin student and a wide teacher. 

A large number of parameters reside in the word embeddings of pre-trained models. For instance, multilingual BERT with WordPiece vocabulary of $V=110K$ tokens and embedding dimension of $D=768$ contains $92MM$ word embedding parameters. We use a dimensionality reduction algorithm, namely, Singular Value Decomposition (SVD) to project the teacher word embeddings of dimension $\mathcal{R}^{|\mathcal{V}| \times d^{\mathcal{T}}}$ to a lower dimensional space $\mathcal{R}^{|\mathcal{V}| \times d^{\mathcal{S}}}$. Given the teacher word embedding matrix of dimension $\mathcal{R}^{|\mathcal{V}| \times d^{\mathcal{T}}}$, 
SVD finds the best $d^{\mathcal{S}}$-dimensional representation that minimizes sum of squares of the projections (of rows) to the subspace. 

\subsection{Hidden Layer Representations} 

The student and teacher models consist of $L^{\mathcal{S}}$ and $L^{\mathcal{T}}$ repeated transformer blocks, where $L^{\mathcal{S}} < L^{\mathcal{T}}$. Considering an input sequence of $n$ tokens $x=\{x_1, x_2, \cdots x_n\}$, the token embedding ${W}$ is added to the position ${PE}$ and segment ${SE}$ embeddings as $z_i(x_i) = W(x_i) + PE(i) + SE(i)$. The input to the network is given by  $\mathcal{H}^0=[z_1, z_2, \cdots z_{|x|}]$. In case of the student, the token embedding is obtained from the SVD-decomposed token embedding of the teacher model as $\mathcal{W}^{\mathcal{S}}$, whereas the position and segment embeddings are learnable embeddings of dimension $d^\mathcal{S}$. 
Transformer blocks repeatedly compute hidden state representations from the output of the previous layer, where hidden states from the $l$th layer of the teacher and student are given by,

{\small
\setlength\abovedisplayskip{0pt}
\setlength\belowdisplayskip{0pt}
\begin{align}
\mathcal{H}^\mathcal{T}_l &= Transformer^\mathcal{T}_l(\mathcal{H}^\mathcal{T}_{l-1}), l \in [1, \cdots \mathcal{L}^\mathcal{T}] \\
    \mathcal{H}^\mathcal{T}_l &= [{h}^\mathcal{T}_{l,1}, {h}^\mathcal{T}_{l,2}, \cdots, {h}^\mathcal{T}_{l,|x|}], l \in [1, 2, \cdots \mathcal{L}^\mathcal{T}] \\
\mathcal{H}^\mathcal{S}_l &= Transformer^\mathcal{S}_l(\mathcal{H}^\mathcal{S}_{l-1}), l \in [1, \cdots \mathcal{L}^\mathcal{S}] \\
\mathcal{H}^\mathcal{S}_l &= [{h}^\mathcal{S}_{l,1}, {h}^\mathcal{S}_{l,2}, \cdots, {h}^\mathcal{S}_{l,|x|}], l \in [1, 2, \cdots \mathcal{L}^\mathcal{S}]
\end{align}
}

\subsection{Multi-head Self-attention} 

Transformers view the input representation as a set of key-value pairs $\{\mathcal{K}, \mathcal{V}\}$ of dimension same as input sequence length $|x|$. Each of the key and values are obtained from hidden state representations of the encoder. Transformers compute the weighted sum of the values, where the weight for each value is obtained by dot-product of the query with the key values as $Attention(\mathcal{Q}, \mathcal{K}, \mathcal{V})=softmax(\frac{\mathcal{Q}\mathcal{K}^T}{\sqrt{n}})\mathcal{V}$. In the context of multi-head attention with several attention heads, the above is computed as follows. Consider $\mathcal{A}_{l, a}, a \in [1, 2, \cdots \mathcal{AH}]$, where $\mathcal{AH}$ is the number of attention heads of the teacher and student. 
Consider the query, key and values obtained by
$\mathcal{Q}_{l,a}=\mathcal{H}_{l-1}\mathcal{W}^\mathcal{Q}_{l,a}$, $\mathcal{K}_{l,a}=\mathcal{H}_{l-1}\mathcal{W}^k_{l,a}$, $\mathcal{V}_{l,a}=\mathcal{H}_{l-1}\mathcal{W}^V_{l,a}$.

Each multi-head attention state of dimension $|x| \times |x|$ from the $l$th layer is given by:

{\small
\setlength\abovedisplayskip{0pt}
\setlength\belowdisplayskip{0pt}
\begin{equation}
\mathcal{A}_{l,a}(\mathcal{Q}_{l,a}, \mathcal{K}_{l,a}, \mathcal{V}_{l,a})=softmax(\frac{\mathcal{Q}_{l,a}\mathcal{K}_{l,a}^T}{\sqrt{n}})\mathcal{V}_{l,a}   
\end{equation}
}

Since our teacher and student are both transformers with similar multi-head attention mechanism, we obtain the corresponding attention states:

{\small
\begin{align}
\mathcal{A}^\mathcal{T}_{l,a},&\ l \in [1, 2, \cdots \mathcal{L}^\mathcal{T}], a \in [1, 2, \cdots \mathcal{AH}]
\\
\mathcal{A}^\mathcal{S}_{l,a},&\ l \in [1, 2, \cdots \mathcal{L}^\mathcal{S}], a \in [1, 2, \cdots \mathcal{AH}]
\end{align}
}


\subsection{Multi-task Multi-layer Distillation}

\noindent{\bf Multi-layer hidden state transfer.} 
We leverage deep representations from multiple layers of the teacher that capture different forms of features to aid the student in learning. In order to align multiple layers of the student to those of the teacher as a form of mimic learning, we train the student with the following multi-layer representation loss objective. Given a deep teacher with $\mathcal{L}^\mathcal{T}$ layers and a shallow student with $\mathcal{L}^\mathcal{S}$ layers, where  $\mathcal{L}^\mathcal{T} > \mathcal{L}^\mathcal{S}$, we align the {\em last}  $\mathcal{L}^\mathcal{S}$ layers of each model. Given a wide teacher and narrow student with corresponding dimensions $d^\mathcal{T} > d^\mathcal{S}$, we perform a linear transformation to upscale and align the corresponding hidden state representations of the student such that $\widetilde{\mathcal{H}}^\mathcal{S}(x)=W^f \cdot \mathcal{H}^\mathcal{S}(x) + b^f$, where $W^f \in R^{d^\mathcal{T} \times d^\mathcal{S}}$ is the transformation matrix, $b^f \in R^{d^\mathcal{T}}$ is the bias.

{\small
\begin{equation}
\setlength\abovedisplayskip{0pt}
\setlength\belowdisplayskip{0pt}
    \label{layer-loss}
    layer_{loss} = -\sum_{
    \substack{l=1 \\ l'=(\mathcal{L}^\mathcal{T}-\mathcal{L}^\mathcal{S})}}^{\substack{l'=\mathcal{L}^\mathcal{T} \\ l=\mathcal{L}^\mathcal{S}}} \sum_{i=1}^{|x|} \frac{|| \widetilde{\mathcal{H}_l}^\mathcal{S}(x_i) - \mathcal{H}_{l'}^\mathcal{T}(x_i) ||^2}{2\cdot\mathcal{L}^\mathcal{S} \cdot |x|}
\end{equation}
}

\noindent {\bf Multi-layer attention transfer.} 
We also leverage the self-attention signals from the different teacher layers to guide the student. Similar to previous loss objective, we align the attention states of the {\em last} $\mathcal{L}^\mathcal{S}$ layers of the teacher and student from multiple attention heads with the following objective:

{\small
\begin{equation}
\setlength\abovedisplayskip{0pt}
\setlength\belowdisplayskip{0pt}
\label{attn-loss}
    attn_{loss} = - \sum_{\substack{l=1 \\ l'={(\mathcal{L}^\mathcal{T}-\mathcal{L}^\mathcal{S})}}}^{\substack{l'=\mathcal{L}^\mathcal{T} \\ l=\mathcal{L}^\mathcal{S}}} \sum_{a=1}^{|\mathcal{AH}|} \sum_{i=1}^{|x|} \frac{|| \mathcal{A}_{l,a}^\mathcal{S}(x_i) - \mathcal{A}_{{l'},a}^\mathcal{T}(x_i) ||^2}{2\cdot\mathcal{L}^\mathcal{S} \cdot |\mathcal{AH}| \cdot |x|} 
\end{equation}
}

\noindent {\bf Task-specific logit transfer.} Given hidden state representations from {\em last layer} $\mathcal{L}^\mathcal{S}$ and $\mathcal{L}^\mathcal{T}$ of the student and teacher, we can obtain the task-specific logits for source transfer task (e.g., MNLI) from:

{\small
\begin{align}
z^\mathcal{S}(x) &= \mathcal{H}_{\mathcal{L}^\mathcal{S}}^\mathcal{S}(x) \cdot W^\mathcal{S}\\
z^\mathcal{T}(x) &= \mathcal{H}_{\mathcal{L}^\mathcal{T}}^\mathcal{T}(x) \cdot W^\mathcal{T}
\end{align}
}

\noindent where $W^{\mathcal{S}} \in R^{d^\mathcal{S} \times C}$, $W^{\mathcal{T}} \in R^{d^\mathcal{T} \times C}$, and $C$ is the number of classes. 
The prior computations of multi-layer hidden state and attention loss are performed over large amounts of unlabeled transfer data from the source task. To explicitly adapt these models to the source task, we leverage some amount of source-task-specific labeled data to align the logits of the teacher and student. To this end, we minimize the following task-specific logit loss:

{\small
\begin{align}
\label{logit-loss}
    logit_{loss} = -\frac{1}{2} || z^{\mathcal{S}}(x) - z^{\mathcal{T}}(x) ||^2
\end{align}
}

Finally, we fine-tune the student on task-specific labeled data with the cross-entropy loss:

{\small
\begin{equation}
\setlength\abovedisplayskip{0pt}
\setlength\belowdisplayskip{0pt}
\label{ce-loss}
    ce_{loss} = - \sum_{c=1}^C \mathcal{I}(x,c)\ log\ softmax(z_c^{\mathcal{S}}(x))
\end{equation}
}

\noindent where $\mathcal{I}(x, c)$ is a binary indicator ($0$ or $1$) if class label $c$ is the correct classification for $x$ and $z_c^{\mathcal{S}}(x)$ is the predicted logit corresponding to class $c$.

\noindent {\bf Progressive knowledge transfer.} Multi-layer joint optimization of the above loss functions bears the risk of error propagation from lower layers impacting the knowledge transfer from upper layers. Recent works~\cite{sun-etal-2020-mobilebert,mukherjee-hassan-awadallah-2020-xtremedistil} demonstrate the benefit of progressive knowledge transfer by gradual freezing and unfreezing of neural network layers for mitigation. We adopt a similar principle in our work with the training recipe in Algorithm~\ref{algo:progressive-transfer}.

Instead of jointly optimizing all the loss functions, we first minimize the multi-layer representation and attention loss to align the last $\mathcal{L}^\mathcal{S}$ layers of the teacher and student. Then we freeze the student encoder, learn task-specific parameters by optimizing task-specific logit loss and cross-entropy loss. For any loss function, we freeze the parameters learned from the previous stage, learn new parameters (e.g., softmax for task-specific loss) introduced by a new loss function, and finally perform end-to-end fine-tuning based on cross-entropy loss. Error propagation from lower layers is mitigated by freezing lower part of the network while learning additional task-specific parameters.

\setlength{\textfloatsep}{0.1cm}
\setlength{\floatsep}{0.1cm}
\begin{algorithm}[t]
\small
{\bf Input:} (a) Transformer teacher model (e.g., BERT) fine-tuned on hard labels on task $\mathcal{S}$ (e.g., MNLI). (b) Initial pre-trained student model (e.g., MiniLM) \; \\
1. Optimize student params $\theta = \{\mathcal{H}_l^{\mathcal{S}} \}_{l=1}^{\mathcal{L}^\mathcal{S}}, \{\mathcal{A}_{l,a}^{\mathcal{S}} \}_{\substack{l=1 \cdots \mathcal{L}^\mathcal{S} \\a=1 \cdots \mathcal{AH} }}$ optimizing losses in Eqn.~\ref{layer-loss} and~\ref{attn-loss} \\
2. Freeze $\theta$ and optimize task-specific parameters $W^{\mathcal{S}}$ optimizing logit loss in Eqn.~\ref{logit-loss} with soft labels \\
3. Update $\theta$ and $W^{\mathcal{S}}$ optimizing logit loss in Eqn.~\ref{logit-loss}  with soft labels \\
4. Freeze $\theta$ and optimize task-specific parameters $W^{\mathcal{S}}$ optimizing cross-entropy loss in Eqn.~\ref{ce-loss} with hard labels \\
5. Update $\theta$ and $W^{\mathcal{S}}$ optimizing cross-entropy loss in Eqn.~\ref{ce-loss} with hard labels \\
 \caption{Progressive knowledge transfer.}
 \label{algo:progressive-transfer}
\end{algorithm}

\section{Experiments}

We first 
explore several augmentation resources for knowledge transfer. Then we compare distillation performance and compression rate of \sysname with existing models on GLUE~\cite{wang-etal-2018-glue}, SQuAD~\cite{rajpurkar-etal-2016-squad} and massive multilingual NER on 41 languages in WikiAnn~\cite{pan-etal-2017-cross}. All experiments are performed on $4$ Tesla V-$100$ GPUs.

\begin{table}[htbp]
  \centering
  \small
  \caption{Performance of \sysname on MNLI with different augmentation resources as unlabeled transfer set with Electra-base-discriminator teacher and pre-initialized MiniLM student (parameters in millions).}
    \begin{tabular}{lrrrr}
    \toprule
    {Model} & Params & Aug. Data & \#Samples & \multicolumn{1}{l}{Acc} \\\midrule
    {\small Electra} & 109 & - & - & 88.12 \\\midrule
    {\small Xtreme} & 22 & MNLI  & 392K&82.56 \\
    {\small DistilTransf.} & & SNLI  & 550K & 82.57 \\
    & & PAWS  & 695K & 82.58 \\
    & & ParaNMT & 5.4MM& 83.70 \\
    & & SentAug & 4.3MM & 84.52 \\\bottomrule
    \end{tabular}%
  \label{tab:mnli-aug}%
\end{table}%

\begin{table*}[htbp]
  \centering
  \small
  \caption{Comparing the performance of distilled models DistilBERT~\cite{sanh2019}, TinyBERT~\cite{jiao2019tinybert}, MiniLM~\cite{wang2020minilm} and \sysname on the development set for several GLUE tasks. R denotes reported published results and HF denotes the performance obtained with our HuggingFace implementations.}
\begin{tabular}{lrrrrrrrrrr}
\toprule
Models &	Params &	Speedup &	MNLI &	QNLI &	QQP &	RTE &	SST	& MRPC &	SQuADv2 &	Avg \\\midrule
    BERT (R) & 109   & 1x     & 84.5  & 91.7  & 91.3  & 68.6  & 93.2  & 87.3  & 76.8  & 84.8 \\
    BERT-Trun (R) & 66    & 2x     & 81.2  & 87.9  & 90.4  & 65.5  & 90.8  & 82.7  & 69.9  & 81.2 \\
    DistilBERT (R) & 66    & 2x     & 82.2  & 89.2  & 88.5  & 59.9  & 91.3  & 87.5  & 70.7  & 81.3 \\
    TinyBERT (R) & 66    & 2x     & 83.5  & 90.5  & 90.6  & 72.2  & 91.6  & 88.4  & 73.1  & 84.3 \\
    MiniLM (R) & 66    & 2x     & 84.0    & 91.0    & 91.0    & 71.5  & 92.0    & 88.4  & 76.4  & 84.9 \\
    MiniLM (R) & 22    & 5.3x   & 82.8  & 90.3  & 90.6  & 68.9  & 91.3  & 86.6  & 72.9  & 83.3 \\\midrule
    BERT (HF) & 109   & 1x     & 84.4  & 91.4 & 91.2 & 66.8 & 93.2 & 83.8 & 74.8  & 83.7 \\
    MiniLM (HF) & 22    & 5.3x   & 82.7 & 89.4 & 90.3  & 64.3 & 90.8 & 84.1 & 71.5  & 81.9 \\
    {\bf XtremeDistilTransf. (HF)} & {\bf 22}    & {\bf 5.3x}   & {\bf 84.5} & {\bf 90.2} & {\bf 90.4} & {\bf 77.3} & {\bf 91.6} & {\bf 89.0}    & {\bf 74.4}  & {\bf 85.3} \\
    XtremeDistilTransf. (HF) & 14    & 9.4x   & 81.8 & 86.9 & 89.5 & 74.4 & 89.9 & 86.5 & 63.0    & 81.7 \\\bottomrule
    \end{tabular}%
  \label{tab:transfer-glue}%
\end{table*}%

\subsection{Unlabeled Augmentation Data}
\label{subsec:aug-data}
Given the best source transfer task as MNLI, we choose the best teacher from Table~\ref{tab:teacher-mnli} as Electra and the best student from Table~\ref{tab:student-glue} as MiniLM for initializing the student model in \sysname. We explore the following augmentation resources; 


\noindent {\bf (a) MNLI} ~\cite{mnli}: We use training data as the transfer set, ignoring the labels.

\noindent {\bf (b) SNLI}~\cite{bowman2015large} is similar to MNLI with human-written English sentence-pairs categorized in three classes (entail / contradict / neutral).

\noindent {\bf (c) PAWS}~\cite{zhang-etal-2019-paws}: This contains human labeled pairs  generated from both word swapping and back translation which feature the importance of modeling structure, context, and word order for identifying paraphrases.

\noindent {\bf (d) ParaNMT}~\cite{wieting-gimpel-2018-paranmt} consists of a large number of English-English sentential paraphrase pairs automatically generated using Neural Machine Translation to translate the Czech side of a large Czech-English parallel corpus.

\noindent {\bf (e) SentAugment}~\cite{du2020selftraining} uses state-of-the-art sentence embeddings to encode the information in a very large bank of sentences from CommonCrawl which can thereafter be used to retrieve in-domain unannotated sentences for any language understanding task. We leverage the SentAugment Sentence Encoder (SASE) with sentencepiece tokenization to encode all the sentences in our MNLI training corpus. Given a pair of NLI sentences $(s_1, s_2)$, we leverage existing code\footnote{https://github.com/facebookresearch/SentAugment} for fast K-nearest neighbor search ($K=10$) to retrieve similar sentences for ${\{s_1^i\}}_{i=1}^K$ and ${\{s_2^j\}}_{j=1}^K$ from pre-trained FAISS\footnote{https://github.com/facebookresearch/faiss} indexes built over the CommonCrawl data. The unlabeled transfer set now consists of all the nearest neighbor pairs $\{s_1^i, s_2^j\}$. 

Table~\ref{tab:mnli-aug} shows the distillation performance with different augmentation resources as unlabeled transfer set. We observe SentAug to perform the best for the MNLI task for two reasons.

\noindent {\bf (a) Amount of transfer data}: Given paraphrasical tasks like PAWS and ParaNMT, we observe ParaNMT to perform better given large-scale transfer data. Similarly, among textual entailment tasks MNLI, SNLI and SentAug (derived from MNLI), SentAug performs better given its scale.

\noindent {\bf (b) Label distribution of source and transfer data.} PAWS and ParaNMT being paraphrasical tasks work with binary labels (e.g., paraphrases or not); whereas, NLI tasks work with ternary labels (e.g., entail, contradict, or neutral). 

\begin{table*}[htbp]
  \centering
  \small
  \caption{Performance comparison on distilling \sysname of varying capacity with teacher models of different sizes. \#TD. and \#Tea. denote \sysname and teacher model parameters in millions.}
    \begin{tabular}{rlrrrrrrrrr}
    \toprule
    \multicolumn{1}{l}{\#TD.} & Teacher & \multicolumn{1}{l}{\#Tea.} & \multicolumn{1}{l}{Speedup} & \multicolumn{1}{l}{MRPC } & \multicolumn{1}{l}{MNLI} & \multicolumn{1}{l}{RTE} & \multicolumn{1}{l}{QQP} & \multicolumn{1}{l}{QNLI} & \multicolumn{1}{l}{SST-2} & \multicolumn{1}{l}{Avg} \\\midrule
    22    & Electra Base & 109   & 5.3x   & 89.0    & 84.5 & 77.3 & 90.4 & 90.2 & 91.6 & 87.2 \\
    22    & BERT Base & 109   & 5.3x   & 88.7 & 84.2 & 75.5 & 90.5 & 90.4 & 92.3 & 86.9 \\
    22    & Electra Large & 335   & 10.6x   & 87.8 & 83.8 & 77.6 & 90.5 & 89.8 & 91.2 & 86.8 \\
    22    & BERT Large & 335   & 10.6x   & 86.5 & 83.3 & 76.9  & 90.5 & 89.7 & 91.1 & 86.3 \\
    14    & Electra Base & 109   & 9.4x    & 86.5 & 81.8 & 74.4 & 89.5 & 86.9 & 89.9 & 84.8 \\\bottomrule
    \end{tabular}%
  \label{tab:teacher-size-eff}%
\end{table*}%

\subsection{Distillation Performance in GLUE}

Given a model distilled from MNLI, we extract the encoder $f(x; \theta_{\mathcal{S}})$, add task-specific parameters $\gamma_{\mathcal{T}}$ and fine-tune $f(x; \theta_{\mathcal{S}}, \gamma_{\mathcal{T}})$ on labeled data for several tasks with results in Table~\ref{tab:transfer-glue}. We report results from both published works and our results built on top of HuggingFace (HF)~\cite{wolf-etal-2020-transformers}.

We observe \sysname to obtain the best performance on an average closely followed by MiniLM and TinyBERT. Since the performance of  pre-trained models vary with careful calibrations, in our implementations with HF {\em default} hyper-parameters (marked as HF in Table~\ref{tab:transfer-glue}), we observe \sysname initialized with MiniLM to outperform MiniLM by $4.3\%$ given same hyper-parameters and random seeds, and outperform BERT by $2\%$ with $5.3x$ inference speedup. In a similar setup, the most compressed version of \sysname with $14MM$ parameters performs within $2.4\%$ of BERT with $9.4x$ inference speedup. 

\subsection{What is more important for distillation, bigger or better teacher?}

Table~\ref{tab:teacher-size-eff} shows the performance of \sysname distilled from teachers of different sizes and pre-training schemes. Refer to Table~\ref{tab:teacher-mnli} for teacher performances on MNLI. We observe that a better teacher (e.g., Electra-base $>$ BERT-base, and Electra-large $>$ BERT-large) leads to a better student.  However, we also observe teacher model complexity to play a significant role in distillation. For instance, although BERT-large is better than Electra-base, we observe a slight degradation in distillation performance when distilled from BERT-large with $3x$ parameters compared to Electra-base --- given the student model of same capacity. We conjecture this to be an artifact of model capacity as it becomes increasingly difficult for a shallow student to mimic a much bigger and deeper teacher.

\subsection{Transfer Distillation for Massive Multilingual Named Entity Recognition}
\label{subsec:language-transfer}

We experiment with \sysname distilled from a monolingual task (MNLI-English) and adapt it for the multilingual setting to perform joint named entity recognition (NER) on $41$ languages.

Consider \sysname encoder with hidden layers $\{\mathcal{H}_l\} \in \mathcal{R}^{|x| \times d}$, attention states $\{\mathcal{A}_{l,a}\} \in \mathcal{R}^{|x| \times |x|}$ and word embeddings $\mathcal{W} \in \mathcal{R}^{|\mathcal{V}| \times d}$, where $|x|, d$ and $|\mathcal{V}|$ denote sequence length, embedding dimension and vocabulary size. The only factor dependent on the vocabulary size is the word embedding matrix $\mathcal{W}$. In principle, we can retain learned hidden layers $\{\mathcal{H}_l\}$ and attention states $\{\mathcal{A}_{l,a}\}$, and only adapt word embeddings to transfer to other languages with different vocabulary. 

To this end, we leverage word embeddings from multilingual BERT for target adaptation.  Specifically, we use word embedding factorization using Singular Value Decomposition (SVD) (as outlined in Section~\ref{subsec:word-emb-factor}) to project $\mathcal{R}^{|\mathcal{V}| \times d^{\mathcal{T}}} \rightarrow \mathcal{R}^{|\mathcal{V}| \times d^{\mathcal{S}}}$ from the mBERT word embedding space to that of \sysname, where $d^{\mathcal{T}}$ and $d^{\mathcal{S}}$ represent the embedding dimension of the teacher and student.

Now, we {\em switch word embedding parameters} in \sysname (distilled from English) with the SVD-decomposed mBERT word embeddings while {\em retaining prior encoder parameters} ($\{\mathcal{H}_l\}, \{\mathcal{A}_{l,a}\}$) and further distil it on the multilingual WikiAnn data from $41$ languages in WikiAnn~\cite{pan-etal-2017-cross}. Table~\ref{tab:multilingual-wikiner} compares the performance of \sysname against multilingual models MMNER~\cite{rahimi-etal-2019-massively} and XtremeDistil~\cite{mukherjee-hassan-awadallah-2020-xtremedistil}. We observe the most compressed version of \sysname with $14$ million parameters to obtain a similar performance to XtremeDistil but with $2x$ additional compression and within $4\%$ F1 of mBERT. Note that, in contrast to XtremeDistil, we transfer encoder parameters from a monolingual distilled model. With progressive knowledge transfer, we further freeze the word embedding parameters and fine-tune the encoder parameters on downstream task.

\begin{table}
  \centering
  \small
  \caption{Massive multilingual NER average F1 scores across $41$ languages. \sysname is distilled from MNLI-EN and adapted for multilingual NER. \#Compres. denotes compression factor with respect to mBert and Params is number of parameters in millions.}
    \begin{tabular}{lrrr}
    \toprule
    Models & \multicolumn{1}{l}{Params} & \multicolumn{1}{l}{\#Compres.} & \multicolumn{1}{l}{F1} \\\midrule
    mBERT-Single & \multicolumn{1}{r}{109*41} & \multicolumn{1}{r}{-} & 90.76 \\
    mBERT & 179   & 1     & 91.86 \\
    MMNER & \multicolumn{1}{r}{28*41} & 0.2   & 89.20 \\
    XtremeDistil & 28    & 6.4   & 88.64 \\
    {\small XtremeDistilTransf.} & 22    & 8.1   & 89.38 \\
    {\small XtremeDistilTransf.} & 14    & 12.8  & 88.18 \\\midrule
    \end{tabular}%
  \label{tab:multilingual-wikiner}%
\end{table}%

\subsection{Ablation Study}
 Table~\ref{tab:ablation} shows ablation results on removing different components from \sysname for multilingual NER. We observe performance degradation on removing multi-layer attention and hidden state losses from distillation objective (a and b). When we remove both multi-layer components in (b), we use hidden-states from only {\em last} layer of the teacher. This also demonstrates the benefit of multi-layer distillation. We observe significant degradation without embedding factorization using SVD (c) i.e. the student uses monolingual (English) word embeddings and vocabulary. Without progressive transfer, and fine-tuning model end-to-end result in some degradation (d). Finally, we observe that distilling a student model from scratch (i.e. randomly initialized) without transferring multilingual word embeddings or encoder parameters (e) result in significant performance loss, thereby, demonstrating the benefit of transfer distillation.
 
 Table 1 in Appendix shows the variation in performance of \sysname with different architecture (number of attention heads, hidden layers and embedding dimension), compression and performance gap against multilingual BERT --- with the smallest version obtaining $87x$ encoder compression (or, $1$ million encoder parameters) with $9\%$ F1 gap against mBERT for NER on $41$ languages. 
 
\begin{table}[htbp]
  \centering
  \small
  \caption{Ablation of \sysname ($22$MM params) on WikiAnn for NER on $41$ languages.}
    \begin{tabular}{lr}
    \toprule
    Distillation Features & {F1}  \\\midrule
    All: w/ multi-layer attn. \& hidden state, &\\ 
    embed. factor. w/ freezing   & 89.38  \\\midrule
    (a) w/o multi-layer attn. & 87.86   \\
    (b) w/o multi-layer attn., w/o hidden state & 87.61   \\
    (c) w/o embed. factor. (monoling. vocab.) & 78.90    \\
    (d) w/ embed. factor \& w/o freezing & 87.76   \\
    (e) init. from scratch & 83.40    \\
    \bottomrule
    \end{tabular}%
  \label{tab:ablation}%
\end{table}%

\section{Related Work}
\noindent{\bf Distillation.} Prior works on task-specific distillation~\citep{DBLP:journals/corr/abs-1904-09482,zhu-etal-2019-panlp,DBLP:journals/corr/abs-1903-12136,turc2019wellread} leverage soft logits from teachers for distilling students.~\cite{sun2019patient,sanh2019,aguilar2019knowledge} leverage teacher representations as additional signals. These methods are often constrained by embedding dimension, width and depth of models. Some recent works leverage embedding~\cite{sun-etal-2020-mobilebert} and shared word~\cite{zhao2019extreme} projection to address these limitations. Task-agnostic methods like~\cite{jiao2019tinybert,wang2020minilm,sun-etal-2020-mobilebert} leverage hidden states and attention states from teachers but not task-specific logits (refer to Table~\ref{tab:contrast-distil-framework} for a contrast). Finally, another line of work in model compression use quantization~\citep{DBLP:journals/corr/GongLYB14}, low-precision training and network pruning~\citep{HanMao16} to reduce the memory footprint.

\noindent{\bf Augmentation}. Unsupervised contrastive learning techniques like SimCLR~\cite{pmlr-v119-chen20j,chen-etal-2020-bridging} leverage semantic equivalence of images to train models to differentiate between images and perturbed versions while obtaining parity with fully-supervised models. Similarly UDA~\cite{xie2019unsupervised} leverages consistency learning between texts and backtranslations to improve few-shot text classification. Finally, self-training and pre-training with SentAugment~\cite{du2020selftraining} improves text classification with task-specific augmentation.

\section{Conclusions}

We develop a novel distillation framework \sysname to leverage the advantages of task-specific distillation for high compression as well as wide applicability of task-agnostic ones. We study transferability of tasks for pre-trained models and demonstrate NLI to be a great source task to obtain a better teacher that transfers well across several tasks. This, in turn, is used to distil a better student obtaining significant improvements over state-of-the-art task-agnostic distilled models over several tasks. Finally, we demonstrate techniques to obtain large-scale task-specific augmentation data from the web to facilitate this knowledge transfer.

\section{Appendix}

\noindent {\bf Hyper-parameters.} \sysname is built over HuggingFace with most of the default hyper-parameters. Please refer to the ReadMe for the attached code for details.

\begin{table*}[htbp]
  \centering
  \small
  \caption{Variation in performance of \sysname with different architecture (in terms of number of hidden layers (\#Layer), attention heads (\#Attn) and hidden dimension (\#Dim) and parameter compression (\#Enc Compress.) with respect to mBERT and corresponding performance gap (F1 Gap\%) for massive multilingual NER on 41 languages. \#Enc. and \#Word Emb. denote the number of encoder and word embedding parameters (in millions, rounded to nearest integer).}
    \begin{tabular}{cccccccc}
    \toprule
        \multicolumn{1}{c}{\# Layer} & \multicolumn{1}{c}{\# Attn} & \multicolumn{1}{c}{\# Dim} & \multicolumn{1}{c}{F1} & \multicolumn{1}{c}{\#Enc.} & \multicolumn{1}{c}{\#Word} & \multicolumn{1}{c}{\#Enc} & \multicolumn{1}{c}{F1 } \\
    \multicolumn{1}{c}{} & \multicolumn{1}{c}{} & \multicolumn{1}{c}{} & \multicolumn{1}{c}{} & \multicolumn{1}{c}{} & \multicolumn{1}{c}{Emb.} & \multicolumn{1}{c}{Compres.} & \multicolumn{1}{c}{Gap(\%)} \\\midrule
    2     & 2     & 128   & 82.05 & 1     & 15    & 87    & 11.49 \\
    6     & 12    & 192   & 83.29 & 8     & 23    & 11    & 10.15 \\
    6     & 12    & 216   & 83.40  & 9     & 26    & 10    & 10.03 \\
    2     & 4     & 256   & 84.30  & 2     & 31    & 44    & 9.06 \\
    4     & 2     & 128   & 84.43 & 1     & 15    & 87    & 8.92 \\
    6     & 2     & 128   & 85.44 & 2     & 15    & 44    & 7.83 \\
    4     & 4     & 256   & 86.26 & 3     & 31    & 29    & 6.95 \\
    4     & 12    & 312   & 86.69 & 5     & 37    & 17    & 6.48 \\
    4     & 12    & 312   & 86.74 & 5     & 37    & 17    & 6.43 \\
    6     & 4     & 256   & 86.82 & 5     & 31    & 17    & 6.34 \\
    6     & 12    & 384   & 88.00    & 11    & 46    & 8     & 5.07 \\
    12    & 12    & 768   & 92.70  & 87    & 92    & 1     & 0 \\\bottomrule
    \end{tabular}%
  \label{tab:transfer-distil-arch}%
\end{table*}

\bibliographystyle{acl_natbib}
\bibliography{anthology,acl2021}

\begin{thebibliography}{41}
\expandafter\ifx\csname natexlab\endcsname\relax\def\natexlab#1{#1}\fi

\bibitem[{Aguilar et~al.(2019)Aguilar, Ling, Zhang, Yao, Fan, and
  Guo}]{aguilar2019knowledge}
Gustavo Aguilar, Yuan Ling, Yu~Zhang, Benjamin Yao, Xing Fan, and Edward Guo.
  2019.
\newblock \href {http://arxiv.org/abs/1910.03723} {Knowledge distillation from
  internal representations}.

\bibitem[{Bowman et~al.(2015)Bowman, Angeli, Potts, and
  Manning}]{bowman2015large}
Samuel~R. Bowman, Gabor Angeli, Christopher Potts, and Christopher~D. Manning.
  2015.
\newblock A large annotated corpus for learning natural language inference.
\newblock In \emph{Proceedings of the 2015 Conference on Empirical Methods in
  Natural Language Processing (EMNLP)}. Association for Computational
  Linguistics.

\bibitem[{Brown et~al.(2020)Brown, Mann, Ryder, Subbiah, Kaplan, Dhariwal,
  Neelakantan, Shyam, Sastry, Askell et~al.}]{brown2020language}
Tom~B Brown, Benjamin Mann, Nick Ryder, Melanie Subbiah, Jared Kaplan, Prafulla
  Dhariwal, Arvind Neelakantan, Pranav Shyam, Girish Sastry, Amanda Askell,
  et~al. 2020.
\newblock Language models are few-shot learners.
\newblock \emph{arXiv preprint arXiv:2005.14165}.

\bibitem[{Chen et~al.(2020{\natexlab{a}})Chen, Frankle, Chang, Liu, Zhang,
  Wang, and Carbin}]{chen2020lottery}
Tianlong Chen, Jonathan Frankle, Shiyu Chang, Sijia Liu, Yang Zhang, Zhangyang
  Wang, and Michael Carbin. 2020{\natexlab{a}}.
\newblock \href {http://arxiv.org/abs/2007.12223} {The lottery ticket
  hypothesis for pre-trained bert networks}.

\bibitem[{Chen et~al.(2020{\natexlab{b}})Chen, Kornblith, Norouzi, and
  Hinton}]{pmlr-v119-chen20j}
Ting Chen, Simon Kornblith, Mohammad Norouzi, and Geoffrey Hinton.
  2020{\natexlab{b}}.
\newblock A simple framework for contrastive learning of visual
  representations.
\newblock In \emph{Proceedings of the 37th International Conference on Machine
  Learning}, volume 119 of \emph{Proceedings of Machine Learning Research},
  pages 1597--1607. PMLR.

\bibitem[{Chen et~al.(2020{\natexlab{c}})Chen, Meng, Li, Chen, Xu, Xu, and
  Zhou}]{chen-etal-2020-bridging}
Xiuyi Chen, Fandong Meng, Peng Li, Feilong Chen, Shuang Xu, Bo~Xu, and Jie
  Zhou. 2020{\natexlab{c}}.
\newblock \href {https://doi.org/10.18653/v1/2020.emnlp-main.275} {Bridging the
  gap between prior and posterior knowledge selection for knowledge-grounded
  dialogue generation}.
\newblock In \emph{Proceedings of the 2020 Conference on Empirical Methods in
  Natural Language Processing (EMNLP)}, pages 3426--3437, Online. Association
  for Computational Linguistics.

\bibitem[{Clark et~al.(2020)Clark, Luong, Le, and
  Manning}]{DBLP:conf/iclr/ClarkLLM20}
Kevin Clark, Minh{-}Thang Luong, Quoc~V. Le, and Christopher~D. Manning. 2020.
\newblock {ELECTRA:} pre-training text encoders as discriminators rather than
  generators.
\newblock In \emph{8th International Conference on Learning Representations,
  {ICLR} 2020, Addis Ababa, Ethiopia, April 26-30, 2020}. OpenReview.net.

\bibitem[{Devlin et~al.(2019{\natexlab{a}})Devlin, Chang, Lee, and
  Toutanova}]{DBLP:conf/naacl/DevlinCLT19}
Jacob Devlin, Ming{-}Wei Chang, Kenton Lee, and Kristina Toutanova.
  2019{\natexlab{a}}.
\newblock {BERT:} pre-training of deep bidirectional transformers for language
  understanding.
\newblock In \emph{Proceedings of the 2019 Conference of the North American
  Chapter of the Association for Computational Linguistics: Human Language
  Technologies, {NAACL-HLT} 2019, Minneapolis, MN, USA, June 2-7, 2019, Volume
  1 (Long and Short Papers)}, pages 4171--4186.

\bibitem[{Devlin et~al.(2019{\natexlab{b}})Devlin, Chang, Lee, and
  Toutanova}]{devlin-etal-2019-bert}
Jacob Devlin, Ming-Wei Chang, Kenton Lee, and Kristina Toutanova.
  2019{\natexlab{b}}.
\newblock \href {https://doi.org/10.18653/v1/N19-1423} {{BERT}: Pre-training of
  deep bidirectional transformers for language understanding}.
\newblock In \emph{Proceedings of the 2019 Conference of the North {A}merican
  Chapter of the Association for Computational Linguistics: Human Language
  Technologies, Volume 1 (Long and Short Papers)}, pages 4171--4186,
  Minneapolis, Minnesota. Association for Computational Linguistics.

\bibitem[{Du et~al.(2020)Du, Grave, Gunel, Chaudhary, Celebi, Auli, Stoyanov,
  and Conneau}]{du2020selftraining}
Jingfei Du, Edouard Grave, Beliz Gunel, Vishrav Chaudhary, Onur Celebi, Michael
  Auli, Ves Stoyanov, and Alexis Conneau. 2020.
\newblock \href {http://arxiv.org/abs/2010.02194} {Self-training improves
  pre-training for natural language understanding}.

\bibitem[{Frankle and Carbin(2019)}]{conf/iclr/FrankleC19}
Jonathan Frankle and Michael Carbin. 2019.
\newblock The lottery ticket hypothesis: Finding sparse, trainable neural
  networks.
\newblock In \emph{ICLR}. OpenReview.net.

\bibitem[{Fu et~al.(2020)Fu, Zhou, Yang, Tang, Liu, Liu, and
  Li}]{fu2020lrcbert}
Hao Fu, Shaojun Zhou, Qihong Yang, Junjie Tang, Guiquan Liu, Kaikui Liu, and
  Xiaolong Li. 2020.
\newblock \href {http://arxiv.org/abs/2012.07335} {Lrc-bert:
  Latent-representation contrastive knowledge distillation for natural language
  understanding}.

\bibitem[{Gong et~al.(2014)Gong, Liu, Yang, and
  Bourdev}]{DBLP:journals/corr/GongLYB14}
Yunchao Gong, Liu Liu, Ming Yang, and Lubomir~D. Bourdev. 2014.
\newblock \href {http://arxiv.org/abs/1412.6115} {Compressing deep
  convolutional networks using vector quantization}.
\newblock \emph{CoRR}, abs/1412.6115.

\bibitem[{Gordon et~al.(2020)Gordon, Duh, and Andrews}]{gordon2020compressing}
Mitchell~A. Gordon, Kevin Duh, and Nicholas Andrews. 2020.
\newblock \href {http://arxiv.org/abs/2002.08307} {Compressing bert: Studying
  the effects of weight pruning on transfer learning}.

\bibitem[{Han et~al.(2016)Han, Mao, and Dally}]{HanMao16}
Song Han, Huizi Mao, and William~J. Dally. 2016.
\newblock Deep compression: Compressing deep neural networks with pruning,
  trained quantization and huffman coding.
\newblock \emph{ICLR}.

\bibitem[{Hinton et~al.(2015)Hinton, Vinyals, and
  Dean}]{DBLP:journals/corr/HintonVD15}
Geoffrey~E. Hinton, Oriol Vinyals, and Jeffrey Dean. 2015.
\newblock \href {http://arxiv.org/abs/1503.02531} {Distilling the knowledge in
  a neural network}.
\newblock \emph{CoRR}, abs/1503.02531.

\bibitem[{Jiao et~al.(2019)Jiao, Yin, Shang, Jiang, Chen, Li, Wang, and
  Liu}]{jiao2019tinybert}
Xiaoqi Jiao, Yichun Yin, Lifeng Shang, Xin Jiang, Xiao Chen, Linlin Li, Fang
  Wang, and Qun Liu. 2019.
\newblock \href {http://arxiv.org/abs/1909.10351} {Tinybert: Distilling bert
  for natural language understanding}.

\bibitem[{Liu et~al.(2019{\natexlab{a}})Liu, He, Chen, and
  Gao}]{DBLP:journals/corr/abs-1904-09482}
Xiaodong Liu, Pengcheng He, Weizhu Chen, and Jianfeng Gao. 2019{\natexlab{a}}.
\newblock \href {http://arxiv.org/abs/1904.09482} {Improving multi-task deep
  neural networks via knowledge distillation for natural language
  understanding}.
\newblock \emph{CoRR}, abs/1904.09482.

\bibitem[{Liu et~al.(2019{\natexlab{b}})Liu, Ott, Goyal, Du, Joshi, Chen, Levy,
  Lewis, Zettlemoyer, and Stoyanov}]{DBLP:journals/corr/abs-1907-11692}
Yinhan Liu, Myle Ott, Naman Goyal, Jingfei Du, Mandar Joshi, Danqi Chen, Omer
  Levy, Mike Lewis, Luke Zettlemoyer, and Veselin Stoyanov. 2019{\natexlab{b}}.
\newblock Roberta: {A} robustly optimized {BERT} pretraining approach.
\newblock \emph{CoRR}, abs/1907.11692.

\bibitem[{Mukherjee and
  Hassan~Awadallah(2020)}]{mukherjee-hassan-awadallah-2020-xtremedistil}
Subhabrata Mukherjee and Ahmed Hassan~Awadallah. 2020.
\newblock \href {https://doi.org/10.18653/v1/2020.acl-main.202}
  {{X}treme{D}istil: Multi-stage distillation for massive multilingual models}.
\newblock In \emph{Proceedings of the 58th Annual Meeting of the Association
  for Computational Linguistics}, pages 2221--2234, Online. Association for
  Computational Linguistics.

\bibitem[{Pan et~al.(2017)Pan, Zhang, May, Nothman, Knight, and
  Ji}]{pan-etal-2017-cross}
Xiaoman Pan, Boliang Zhang, Jonathan May, Joel Nothman, Kevin Knight, and Heng
  Ji. 2017.
\newblock \href {https://doi.org/10.18653/v1/P17-1178} {Cross-lingual name
  tagging and linking for 282 languages}.
\newblock In \emph{Proceedings of the 55th Annual Meeting of the Association
  for Computational Linguistics (Volume 1: Long Papers)}, pages 1946--1958,
  Vancouver, Canada. Association for Computational Linguistics.

\bibitem[{Raffel et~al.(2019)Raffel, Shazeer, Roberts, Lee, Narang, Matena,
  Zhou, Li, and Liu}]{Raffel2019ExploringTL}
Colin Raffel, Noam Shazeer, Adam Roberts, Katherine Lee, Sharan Narang, Michael
  Matena, Yanqi Zhou, Wei Li, and Peter~J. Liu. 2019.
\newblock Exploring the limits of transfer learning with a unified text-to-text
  transformer.
\newblock \emph{ArXiv}, abs/1910.10683.

\bibitem[{Rahimi et~al.(2019)Rahimi, Li, and Cohn}]{rahimi-etal-2019-massively}
Afshin Rahimi, Yuan Li, and Trevor Cohn. 2019.
\newblock \href {https://doi.org/10.18653/v1/P19-1015} {Massively multilingual
  transfer for {NER}}.
\newblock In \emph{Proceedings of the 57th Annual Meeting of the Association
  for Computational Linguistics}, pages 151--164, Florence, Italy. Association
  for Computational Linguistics.

\bibitem[{Rajpurkar et~al.(2016)Rajpurkar, Zhang, Lopyrev, and
  Liang}]{rajpurkar-etal-2016-squad}
Pranav Rajpurkar, Jian Zhang, Konstantin Lopyrev, and Percy Liang. 2016.
\newblock \href {https://doi.org/10.18653/v1/D16-1264} {{SQ}u{AD}: 100,000+
  questions for machine comprehension of text}.
\newblock In \emph{Proceedings of the 2016 Conference on Empirical Methods in
  Natural Language Processing}, pages 2383--2392, Austin, Texas. Association
  for Computational Linguistics.

\bibitem[{Sanh(2019)}]{sanh2019}
Victor Sanh. 2019.
\newblock Introducing distilbert, a distilled version of bert.
\newblock \url{https://medium.com/huggingface/distilbert-8cf3380435b5}.

\bibitem[{Strubell et~al.(2019)Strubell, Ganesh, and
  McCallum}]{strubell-etal-2019-energy}
Emma Strubell, Ananya Ganesh, and Andrew McCallum. 2019.
\newblock \href {https://doi.org/10.18653/v1/P19-1355} {Energy and policy
  considerations for deep learning in {NLP}}.
\newblock In \emph{Proceedings of the 57th Annual Meeting of the Association
  for Computational Linguistics}, pages 3645--3650, Florence, Italy.
  Association for Computational Linguistics.

\bibitem[{Sun et~al.(2019)Sun, Cheng, Gan, and Liu}]{sun2019patient}
Siqi Sun, Yu~Cheng, Zhe Gan, and Jingjing Liu. 2019.
\newblock \href {http://arxiv.org/abs/1908.09355} {Patient knowledge
  distillation for bert model compression}.

\bibitem[{Sun et~al.(2020)Sun, Yu, Song, Liu, Yang, and
  Zhou}]{sun-etal-2020-mobilebert}
Zhiqing Sun, Hongkun Yu, Xiaodan Song, Renjie Liu, Yiming Yang, and Denny Zhou.
  2020.
\newblock \href {https://doi.org/10.18653/v1/2020.acl-main.195}
  {{M}obile{BERT}: a compact task-agnostic {BERT} for resource-limited
  devices}.
\newblock In \emph{Proceedings of the 58th Annual Meeting of the Association
  for Computational Linguistics}, pages 2158--2170, Online. Association for
  Computational Linguistics.

\bibitem[{Tang et~al.(2019)Tang, Lu, Liu, Mou, Vechtomova, and
  Lin}]{DBLP:journals/corr/abs-1903-12136}
Raphael Tang, Yao Lu, Linqing Liu, Lili Mou, Olga Vechtomova, and Jimmy Lin.
  2019.
\newblock \href {http://arxiv.org/abs/1903.12136} {Distilling task-specific
  knowledge from {BERT} into simple neural networks}.
\newblock \emph{CoRR}, abs/1903.12136.

\bibitem[{Turc et~al.(2019)Turc, Chang, Lee, and Toutanova}]{turc2019wellread}
Iulia Turc, Ming-Wei Chang, Kenton Lee, and Kristina Toutanova. 2019.
\newblock \href {http://arxiv.org/abs/1908.08962} {Well-read students learn
  better: On the importance of pre-training compact models}.

\bibitem[{Wang et~al.(2018)Wang, Singh, Michael, Hill, Levy, and
  Bowman}]{wang-etal-2018-glue}
Alex Wang, Amanpreet Singh, Julian Michael, Felix Hill, Omer Levy, and Samuel
  Bowman. 2018.
\newblock \href {https://doi.org/10.18653/v1/W18-5446} {{GLUE}: A multi-task
  benchmark and analysis platform for natural language understanding}.
\newblock In \emph{Proceedings of the 2018 {EMNLP} Workshop {B}lackbox{NLP}:
  Analyzing and Interpreting Neural Networks for {NLP}}, pages 353--355,
  Brussels, Belgium. Association for Computational Linguistics.

\bibitem[{Wang et~al.(2020)Wang, Wei, Dong, Bao, Yang, and
  Zhou}]{wang2020minilm}
Wenhui Wang, Furu Wei, Li~Dong, Hangbo Bao, Nan Yang, and Ming Zhou. 2020.
\newblock \href {http://arxiv.org/abs/2002.10957} {Minilm: Deep self-attention
  distillation for task-agnostic compression of pre-trained transformers}.

\bibitem[{Wieting and Gimpel(2018)}]{wieting-gimpel-2018-paranmt}
John Wieting and Kevin Gimpel. 2018.
\newblock \href {https://doi.org/10.18653/v1/P18-1042} {{P}ara{NMT}-50{M}:
  Pushing the limits of paraphrastic sentence embeddings with millions of
  machine translations}.
\newblock In \emph{Proceedings of the 56th Annual Meeting of the Association
  for Computational Linguistics (Volume 1: Long Papers)}, pages 451--462,
  Melbourne, Australia. Association for Computational Linguistics.

\bibitem[{Williams et~al.(2018)Williams, Nangia, and Bowman}]{mnli}
Adina Williams, Nikita Nangia, and Samuel Bowman. 2018.
\newblock A broad-coverage challenge corpus for sentence understanding through
  inference.
\newblock In \emph{Proceedings of the 2018 Conference of the North American
  Chapter of the Association for Computational Linguistics: Human Language
  Technologies, Volume 1 (Long Papers)}, pages 1112--1122. Association for
  Computational Linguistics.

\bibitem[{Wolf et~al.(2020)Wolf, Debut, Sanh, Chaumond, Delangue, Moi, Cistac,
  Rault, Louf, Funtowicz, Davison, Shleifer, von Platen, Ma, Jernite, Plu, Xu,
  Le~Scao, Gugger, Drame, Lhoest, and Rush}]{wolf-etal-2020-transformers}
Thomas Wolf, Lysandre Debut, Victor Sanh, Julien Chaumond, Clement Delangue,
  Anthony Moi, Pierric Cistac, Tim Rault, Remi Louf, Morgan Funtowicz, Joe
  Davison, Sam Shleifer, Patrick von Platen, Clara Ma, Yacine Jernite, Julien
  Plu, Canwen Xu, Teven Le~Scao, Sylvain Gugger, Mariama Drame, Quentin Lhoest,
  and Alexander Rush. 2020.
\newblock \href {https://doi.org/10.18653/v1/2020.emnlp-demos.6} {Transformers:
  State-of-the-art natural language processing}.
\newblock In \emph{Proceedings of the 2020 Conference on Empirical Methods in
  Natural Language Processing: System Demonstrations}, pages 38--45, Online.
  Association for Computational Linguistics.

\bibitem[{Wu et~al.(2016)Wu, Schuster, Chen, and
  et~al.}]{DBLP:journals/corr/WuSCLNMKCGMKSJL16}
Yonghui Wu, Mike Schuster, Zhifeng Chen, and Quoc V.~Le et~al. 2016.
\newblock \href {http://arxiv.org/abs/1609.08144} {Google's neural machine
  translation system: Bridging the gap between human and machine translation}.
\newblock \emph{CoRR}, abs/1609.08144.

\bibitem[{Xie et~al.(2019)Xie, Dai, Hovy, Luong, and Le}]{xie2019unsupervised}
Qizhe Xie, Zihang Dai, Eduard Hovy, Minh-Thang Luong, and Quoc~V. Le. 2019.
\newblock \href {http://arxiv.org/abs/1904.12848} {Unsupervised data
  augmentation for consistency training}.

\bibitem[{Yang et~al.(2019)Yang, Dai, Yang, Carbonell, Salakhutdinov, and
  Le}]{DBLP:journals/corr/abs-1906-08237}
Zhilin Yang, Zihang Dai, Yiming Yang, Jaime~G. Carbonell, Ruslan Salakhutdinov,
  and Quoc~V. Le. 2019.
\newblock \href {http://arxiv.org/abs/1906.08237} {Xlnet: Generalized
  autoregressive pretraining for language understanding}.
\newblock \emph{CoRR}, abs/1906.08237.

\bibitem[{Zhang et~al.(2019)Zhang, Baldridge, and He}]{zhang-etal-2019-paws}
Yuan Zhang, Jason Baldridge, and Luheng He. 2019.
\newblock \href {https://doi.org/10.18653/v1/N19-1131} {{PAWS}: Paraphrase
  adversaries from word scrambling}.
\newblock In \emph{Proceedings of the 2019 Conference of the North {A}merican
  Chapter of the Association for Computational Linguistics: Human Language
  Technologies, Volume 1 (Long and Short Papers)}, pages 1298--1308,
  Minneapolis, Minnesota. Association for Computational Linguistics.

\bibitem[{Zhao et~al.(2019)Zhao, Gupta, Song, and Zhou}]{zhao2019extreme}
Sanqiang Zhao, Raghav Gupta, Yang Song, and Denny Zhou. 2019.
\newblock \href {http://arxiv.org/abs/1909.11687} {Extreme language model
  compression with optimal subwords and shared projections}.

\bibitem[{Zhu et~al.(2019)Zhu, Zhou, Wang, Luo, Li, Ni, and
  Xie}]{zhu-etal-2019-panlp}
Wei Zhu, Xiaofeng Zhou, Keqiang Wang, Xun Luo, Xiepeng Li, Yuan Ni, and Guotong
  Xie. 2019.
\newblock \href {https://doi.org/10.18653/v1/W19-5040} {{PANLP} at {MEDIQA}
  2019: Pre-trained language models, transfer learning and knowledge
  distillation}.
\newblock In \emph{Proceedings of the 18th BioNLP Workshop and Shared Task},
  pages 380--388, Florence, Italy. Association for Computational Linguistics.

\end{thebibliography}

\end{document}